\documentclass[11pt]{article}

\usepackage[preprint]{acl}

\usepackage{times}
\usepackage{latexsym}
\usepackage{booktabs}
\usepackage{amsmath} 
\usepackage{amssymb}
\usepackage{multirow}
\usepackage{multirow}
\usepackage{newunicodechar}
\newunicodechar{θ}{\theta}
\usepackage{xspace}

\newunicodechar{♫}{\textmusicalnote}
\newunicodechar{∼}{\textasciitilde}
\newunicodechar{∗}{\ast}
\usepackage[T1]{fontenc}

\usepackage{graphicx}
\usepackage[utf8]{inputenc}
\usepackage{tabularx}
\usepackage{booktabs}
\usepackage{array}

\newcolumntype{Y}{>{\centering\arraybackslash}X}

\usepackage{microtype}

\usepackage{inconsolata}

\usepackage{graphicx}

\newcommand{\method}{RASER\xspace}
\newcommand{\methodtwo}{RASER-2\xspace}
\newcommand{\methodthree}{RASER-3\xspace}
\newcommand{\stoproute}{\textsc{One-shot RAG}\xspace}
\newcommand{\pruneroute}{\textsc{Prune}\xspace}
\newcommand{\iterroute}{\textsc{IRCoT}\ensuremath{^{*}}\xspace}
\newcommand{\decomproute}{\textsc{Self-Ask}\ensuremath{^{*}}\xspace}
\newcommand{\fone}{F\textsubscript{1}\xspace}

\title{RASER: Recoverability-Aware Selective Escalation Router for Multi-Hop Question Answering}

\author{
Yuyang Li$^{1}$ \quad Zihe Yan$^{2}$ \quad Tobias K{\"a}fer$^{1}$ \\
$^{1}$Institute AIFB, Karlsruhe Institute of Technology, Karlsruhe, Germany \\
$^{2}$Shanghai Jiao Tong University, Shanghai, China \\
\texttt{\{yuyang.li,tobias.kaefer\}@kit.edu} \\
\texttt{yangtuomao@sjtu.edu.cn}
}

\definecolor{stopfill}{RGB}{255,247,219}      
\definecolor{stopframe}{RGB}{200,170,80}
\definecolor{prunefill}{RGB}{255,229,229}     
\definecolor{pruneframe}{RGB}{200,90,90}
\definecolor{iterfill}{RGB}{222,236,255}      
\definecolor{iterframe}{RGB}{70,110,180}
\definecolor{stepfill}{RGB}{242,242,242}      
\definecolor{goodgreen}{RGB}{198,232,180}
\definecolor{badred}{RGB}{248,205,205}
\definecolor{promptframe}{RGB}{120,120,120}
\definecolor{prompttitle}{RGB}{225,232,242}

\newcommand{\promptbox}[2]{%
  \par\vspace{4pt}\noindent
  \setlength{\fboxsep}{4pt}\setlength{\fboxrule}{0.5pt}%
  \fcolorbox{promptframe}{white}{%
    \begin{minipage}{\dimexpr 0.94\linewidth-2\fboxsep\relax}
      \colorbox{prompttitle}{\parbox{\linewidth}{\strut\textbf{#1}\strut}}\par
      \vspace{4pt}\small #2
    \end{minipage}%
  }%
  \par\vspace{6pt}
}

\begin{document}
\maketitle
\begin{abstract}
Multi-hop question-answering systems often use expensive retrieval on every question. They may decompose the question, run several retrieval rounds, or search through bridge entities before answering. All of these strategies rely on repeated LLM calls to rewrite or decompose the question, which increases extra token cost, and it is not fitting when the LLM budget is tight. However, our analysis shows that lots of multi-hop questions are already answered correctly by a single one-shot RAG, so running an extra retrieval on every question wastes the budget. We introduce \textbf{RASER} (\textbf{R}ecoverability-\textbf{A}ware \textbf{S}elective \textbf{E}scalation \textbf{R}outer), a family of cheap routers built on one-shot RAG and six features from it. RASER-2 decides whether to stop or escalate to the extra-retrieval action PRUNE. RASER-3 chooses among one-shot RAG, PRUNE, and iterative retrieval IRCoT, using the same features but adding an explicit cost-accuracy trade-off. Neither router makes an extra LLM call to decide. Across six LLMs and three multi-hop QA benchmarks, both routers stay competitive with the other state-of-the-art (SOTA) baselines in \fone while spending only 41–49\% of always-prune's tokens and also less than the iterative and decomposition retrieval baselines. Code is available at \url{https://github.com/YuyangLi99/RASER_2026}.

\end{abstract}

\section{Introduction}
\label{sec:intro}

Multi-hop question answering (QA) requires a system to integrate evidence, which are retrieved passages that support the answer from multiple texts, to produce an answer. A passage is a text chunk returned by the retriever. For example, answering a question such as "Who is the spouse of the director of \textit{Inception}?" requires first identifying the director of \textit{Inception} and then retrieving evidence about that person's spouse. This makes multi-hop QA different from single-hop QA: the difficulty is not only finding a relevant passage but also deciding which intermediate entity or fact should guide the next retrieval step. Benchmarks such as HotpotQA, 2WikiMultiHopQA, and MuSiQue were introduced to evaluate this kind of multi-step reasoning, with MuSiQue in particular designed to reduce shortcut solutions and require connected reasoning \citep{yang-etal-2018-hotpotqa, ho-etal-2020-constructing, trivedi-etal-2022-musique}.

In a retrieval-augmented generation (RAG) system, the answer is produced by a large language model (LLM) after retrieving passages. And a common solution to this challenge is \emph{multi-step retrieval}: instead of retrieving once and answering immediately, the system repeatedly expands the retrieved passages. These methods are effective because they can recover evidence that a one-shot retriever misses. However, they also apply an extra retrieval for every question: every question pays the cost of extra retrieval, extra LLM calls, or query rewriting, even when the one-shot answer is already correct or when no tested retrieval strategy can recover the answer.

This paper asks: \emph{When is additional retrieval actually necessary?} To answer it, we perform a recoverability analysis over multiple retrieval strategies. For each question, we run a one-shot dense retrieval path and several bridge-style retrieval paths and then compare their answer \fone after the fact. This analysis measures the potential benefit if a router always chose the best strategy for each question. The analysis reveals only a minority of questions improve by at least 0.1 F1 from bridge-style retrieval, while many are already solved by one-shot retrieval, and another 14--27\% remains unrecoverable under all tested strategies. This suggests that the main challenge in multi-hop QA is not simply how to retrieve more evidence but how to decide \emph{when} bridge retrieval is worth paying for.

We therefore define multi-hop QA as a \textbf{recoverability-aware selective escalation} problem. Rather than always applying iterative retrieval, our method first produces a one-shot answer from dense retrieval, then predicts whether that answer is sufficiently supported by the retrieved evidence or whether the question should be escalated to a bridged retrieval path. This framing is related to prior adaptive retrieval and routing work \citep{jeong-etal-2024-adaptive,guo-etal-2026-deepsieve,liu2026a2ragadaptiveagenticgraph}, but differs in two important respects. First, the decision target is \emph{recoverability}: will bridge retrieval help this question under the current retrieved passages? Second, the router is implemented as \textbf{a lightweight Gradient Boosting Machine (GBM)-based classifier} over cheap answer-side, retrieval-side, and evidence-support features, introducing zero router-side LLM tokens.

Our method, \textbf{RASER} (\textbf{R}ecoverability-\textbf{A}ware \textbf{S}elective \textbf{E}scalation \textbf{R}outer), is a family of routers built on the same one-shot RAG and six features. The first one-shot RAG returns the first draft answer and top-k chunks, and then it produces the six features: retrieval similarity scores, draft-answer length, question type, and bridge gaps. RASER-2 is a 2-action router; it is a gradient-boosted binary classifier over six features to pick between one-shot RAG and an extra retrieval route, "PRUNE," under a fixed threshold. RASER-3 is a 3-action cost-aware router. It uses the same six features to predict the answer quality of one-shot RAG, PRUNE, and an iterative retrieval route, "IRCoT," and then chooses the best route based on a token-cost penalty. Both routers make their decision without an extra LLM call. Under a unified setup with the same retriever, LLMs, corpus, and chunking text across all baselines, RASER achieves \emph{competitive F1 at a substantially lower token cost} than strong iterative baselines. Our paper has the following contributions:

\begin{itemize}
    \item We formulate multi-hop QA as a \textbf{recoverability-aware selective escalation} problem and show through recoverability analysis that only a minority of questions improve by at least 0.1 F1 from additional retrieval.
    \item We propose \textbf{RASER}, a family of lightweight routers that includes RASER-2 and RASER-3. They use zero router-side LLM tokens and cheap evidence-support features.
    \item We provide a \textbf{unified comparison} among all the baselines and a route-wise analysis showing that RASER attains competitive \fone at much lower token cost overall.
\end{itemize}

\section{Related Work}
\label{sec:related work}

\paragraph{Multi-hop QA Benchmarks}

Multi-hop QA datasets move beyond single-evidence question answering by requiring systems to connect entities, facts, or supporting passages. HotpotQA \citep{yang-etal-2018-hotpotqa} introduced diverse, explainable multi-hop QA with supporting facts and remains the most widely used benchmark in this area. 2WikiMultiHopQA \citep{ho-etal-2020-constructing} was designed to ensure multi-hop reasoning through structured generation over Wikidata and Wikipedia, while also exposing reasoning paths and supporting evidence. MuSiQue \citep{trivedi-etal-2022-musique} was proposed specifically to reduce shortcut solutions by composing connected single-hop questions, making it a stronger testbed for genuine multi-hop reasoning. 

\paragraph{Extra Retrieval for Multi-hop QA}

A large part of previous work improves multi-hop QA  by adding retrieval or reasoning steps. IRCoT interleaves chain-of-thought reasoning with retrieval queries \cite{trivedi-etal-2023-interleaving}, and ITER-RetGen alternates between retrieval and generation in the document \cite{shao-etal-2023-enhancing}. Decomposition approaches such as Self-Ask, Least-to-Most Prompting, and Decomposed Prompting make intermediate questions explicit\cite{press-etal-2023-measuring, zhou2023leasttomost, khot2023decomposedpromptingmodularapproach}. In the recent work, ChainRAG \citep{zhu-etal-2025-mitigating} mitigates lost-in-retrieval failures through progressive retrieval, sub-question answering, and query rewriting. KiRAG \citep{fang-etal-2025-kirag} instead performs knowledge-driven iterative retrieval by decomposing documents into triples and dynamically retrieving bridge knowledge that fills information gaps. Both methods show that one-shot retrieval is often insufficient for multi-hop reasoning.

\paragraph{Adaptive Retrieval}

Several systems have already avoided using the same retrieval method for every input. Adaptive-RAG \citep{jeong-etal-2024-adaptive} sees retrieval as a complexity-based decision and learns to select among no-retrieval, single-step retrieval, and iterative retrieval strategies. DeepSieve \citep{guo-etal-2026-deepsieve} uses an LLM as a knowledge router across heterogeneous knowledge sources; FLARE retrieves when a forward-looking generation appears uncertain \cite{jiang-etal-2023-active}, and DRAGIN triggers retrieval when token-level signals suggest the model needs outside information \cite{su-etal-2024-dragin}.

\paragraph{Selective prediction and cost control}

The closest idea is selective prediction, where a model may abstain from examples likely to be wrong \cite{NIPS2017_4a8423d5}. The system still answers every question; it only abstains from paying for a more expensive retrieval action when that action is unlikely to change the answer. With a 3-action router, the same idea becomes action selection: if paying more is worthy, the system must also decide whether a bridge or an iterative action is the better use of the budget.

\section{Recoverability Analysis}
\label{sec:oracle}

Before introducing RASER, we first ask a diagnostic question: \emph{how many multi-hop questions actually need additional retrieval?} Although multi-hop QA benchmarks are designed to require reasoning over multiple pieces of evidence, not every question necessarily benefits from running an expensive iterative retrieval pipeline. Some questions are already answered correctly by one-shot retrieval, while others remain difficult even after additional retrieval.
To quantify this, we perform a per-sample oracle analysis over three representative strategies: \textsc{one-shot RAG}, Prune, IRCoT, and \textsc{SELF-ASK}.

For each question, we run all strategies and compute the answer F1 produced by each one. It tells us how the dataset decomposes into questions that are already solved, questions that can be helped by bridge retrieval, and questions that none of the tested methods can recover.

\begin{table}[t]
\centering
\scriptsize
\setlength{\tabcolsep}{3pt}
\renewcommand{\arraystretch}{1.08}

\begin{tabularx}{\linewidth}{@{}lYYYY@{}}
\toprule
Reader 
& \shortstack{one-shot RAG\\correct}
& \shortstack{no method\\helps}
& \shortstack{small /\\unclear gain}
& \shortstack{bridge\\helps}\\
\midrule
GPT-OSS-120B    & \textbf{53\%} & 14\% & 19\% & 14\%\\
Gemma-3-31B     & \textbf{48\%} & 19\% & 13\% & 20\%\\
Mistral-S-119B  & 37\%          & 20\% &  9\% & \textbf{34\%}\\
Phi-4-mini      & 31\%          & 25\% & 15\% & 29\%\\
Llama-3.1-8B    & 27\%          & 27\% &  8\% & \textbf{38\%}\\
Llama-3-8B      & \textbf{24\%} & 26\% & 12\% & \textbf{37\%}\\
\bottomrule
\end{tabularx}
\caption{What happens to each question if we try deeper retrieval? We assign a question by the best result among three expensive bridge methods: \pruneroute, iterative \iterroute, and decomposition \decomproute. The first two columns are questions that should \emph{not} be escalated: either the one-shot answer is already correct, or no methods fix it. The last column is a clear target for bridging retrieval: questions where \emph{some} expensive bridges improve the answer by $\ge 0.1$ \fone more than a one-shot retrieval.}
\label{tab:oracle_decomposition}
\end{table}

Table~\ref{tab:oracle_decomposition} shows that lots of questions do not clearly benefit from extra retrieval. This recoverability analysis reframes the task: the central problem is not \emph{how to always retrieve more}, but \emph{how to detect when bridge retrieval is worth paying for}.

\section{Approach}
\label{Approach}

\begin{figure*}[t]
\centering
\includegraphics[width=0.88\textwidth]{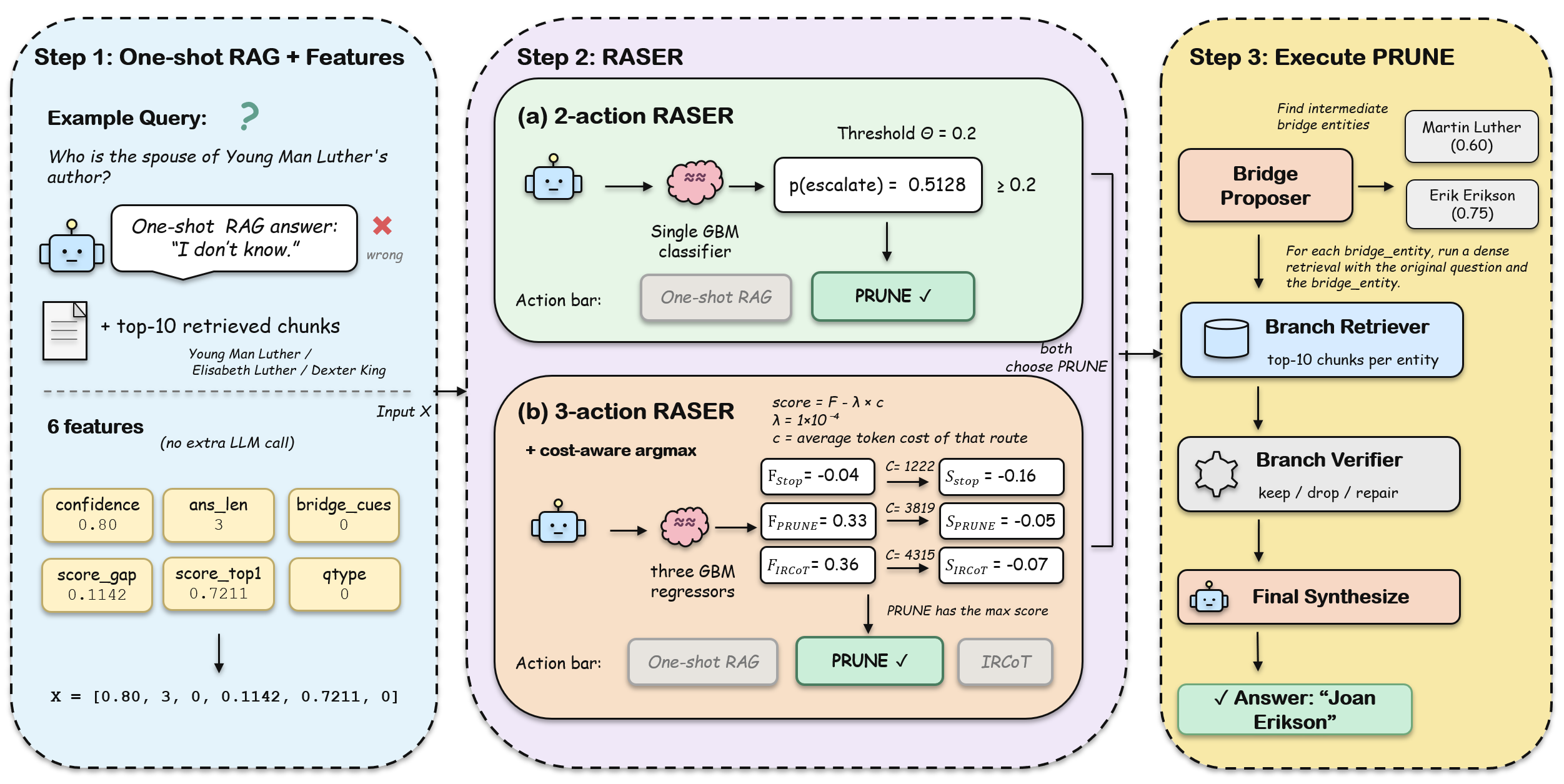}
\caption{\textbf{\method workflow.} The example question needs two reasoning hops to answer (the author of \emph{Young Man Luther} is Erik Erikson, whose spouse is Joan Erikson), and the one-shot RAG gets the wrong answer (\emph{"I don't know"}). It also produces six features. RASER reads those same six features to decide which action to choose. The \textbf{2-action RASER} (a) is a single GBM classifier: it picks between stopping directly after one-shot RAG and PRUNE by checking bridge entities and doing extra retrieval. The \textbf{3-action RASER} (b) is three predictors plus a token-cost penalty: it picks among one-shot RAG, \pruneroute, and \iterroute by weighing predicted accuracy against the tokens each action would spend. For this question, both routers choose PRUNE. Step~3 shows the Bridge Proposer proposing two bridge candidates (\emph{Martin Luther} and \emph{Erik Erikson}); the Branch Retriever re-retrieves chunks for each, the Branch Verifier keeps the useful ones, and the Final Synthesizer combines the bridges with the chunks to return the correct answer \emph{Joan Erikson}. Prompts are in Appendix \ref{app:prompts}.}
\label{fig:raser_mechanism}
\end{figure*}

Section \ref{sec:oracle} shows that extra retrieval is useful only for some questions and differs by LLMs and datasets. But running extra retrieval on every question wastes budget, because lots of questions do not need it. What we want instead is a way to spend the expensive budget where it pays off. This section describes two routers. Our method, \textbf{RASER} (\textbf{R}ecoverability-\textbf{A}ware \textbf{S}elective \textbf{E}scalation \textbf{R}outer), is not a new retriever or a new LLM. It is a small decision layer after the first one-shot retrieval-and-answer. It is used to answer one practical question: Return the one-shot retrieved answer, or spend more tokens on one of the extra retrieval routes and answer again? The 2-action RASER is the low-budget router; it decides whether to stop or run one bridge step. The 3-action RASER is the cost-aware router: it decides whether to stop, bridge once, or run iterative retrieval. Figure \ref{fig:raser_mechanism} shows one example question through the whole pipeline; the rest of this section
will explain each part.

\subsection{2-action RASER: two-route recoverability classifier}


A 2-action RASER chooses between two actions. The cheap action, one-shot RAG, just retrieves once and asks the LLM for an answer. The expensive action, PRUNE, tries to find a bridge entity, which is the missing entity that links the question's two hops. For example, to answer, "Who is the spouse of \emph{Young Man Luther}'s author?" You first have to identify the bridge entity, \emph{Erik Erikson} (the author of the book). Then you can look up his spouse with this bridge entity. PRUNE has four steps: ask the LLM to propose up to two candidate bridge entities, re-retrieve using each one, drop weak entities with a lightweight rule-based verifier, and ask the reader for a final answer. 

2-action RASER runs four steps: 1. Run one-shot RAG. Retrieve the top-k chunks once and ask the LLM for a draft answer. 2. Summarize the first RAG results. Build a six-feature vector from the retrieved scores, the draft answer, and the question text. 3. Predict whether extra retrieval will help. A GBM (Gradient Boosting Machine) classifier estimates p(BRIDGEABLE | s), the probability that a PRUNE action will improve the draft answer. 4. Choose an action. If the probability is at least Threshold $\theta$, run the bridge action PRUNE and return its answer; otherwise, return the draft answer from one-shot RAG. Appendix~\ref{app:sensitivity} shows why we select $\theta$ = 0.20. RASER is trained by six features (Table \ref{tab:features}). Two features describe the draft answer, two features describe the retrieval scores, one feature looks for bridge-like words in the question, and one feature records the question type. All of them are available after one-shot RAG, so adding the router does not require any language model calls or retrieval methods.

For the \emph{Young Man Luther} question above (Figure~\ref{fig:raser_mechanism}), one-shot RAG answers \emph{"I don't know"}, which is not the gold answer \emph{Joan Erikson}. The first-shot RAG also gives the router useful signals: the answer is short and an "I Dont know"(\texttt{ans\_len}\,$=3$, \texttt{confidence}\,$=0$); the top retrieved chunks are about Luther family members but not the name of the author of the book (\texttt{score\_gap}\,$=0.1142$, no chunk clearly wins); and the question asks for a person (\texttt{qtype}\,$=$\,entity). From these, the classifier assigns $p(\textsc{Bridgeable})=0.5128$, above the threshold $\theta=0.20$, so \methodtwo runs \pruneroute. The Bridge Proposer extracts \emph{Erik Erikson} as the missing fact, the Branch Retriever pulls Erikson-related chunks, and the Final Synthesizer returns \emph{Joan Erikson}. The details of the prompts are shown in the Appendix \ref{app:prompts}.

\begin{table}[t]
\centering
\small
\setlength{\tabcolsep}{4pt}
\begin{tabular}{p{0.28\linewidth}p{0.6\linewidth}}
\toprule
Feature & Definition\\
\midrule
\texttt{confidence} & Rule-based confidence score from one-shot RAG and the question type.\\
\texttt{ans\_len}   & Number of words in the draft answer.\\
\texttt{bridge\_cues} & Did the question text match any multi-hop linguistic pattern?
 e.g.,\
\emph{"X of the Y of Z"}, \emph{"who ...was born/lived/located in ...where/which/what,"}  or 
                        \emph{"the country where"}.\\
\texttt{score\_gap} & Gap between the top-1 and top-5 dense
                      retrieval scores. Big gap = clear winner; tiny gap = ambiguous.\\
\texttt{score\_top1} & Top-1 dense retrieval score. Cosine similarity between the question and the top-1 retrieved chunk \\
\texttt{qtype}      & Question type (entity = 0 / date = 1 / yes-no = 2 / count = 3 /
                      other = 4), assigned by simple rules.\\
\bottomrule
\end{tabular}
\caption{The six input features to the RASER. They come after the
one-shot RAG: the draft answer, the top-10 retrieved chunks' similarity scores, and the question itself. Computing them doesn't require extra LLM calls.}
\label{tab:features}
\end{table}

\subsection{3-action RASER: three-route cost-aware classifier}

The 2-action RASER answers only, "Should we escalate?" That works as long as there is one obvious expensive action to escalate. But there isn't. Which expensive route helps change with the LLM? PRUNE wins on some of them and IRCOT* on others. A 3-action router handles that by extending the routes and adding a price tag. The routes grow to three choices: one-shot RAG, PRUNE, and the iterative route IRCOT*—and the router learns with each question \emph{whether} to escalate and \emph{which} expensive route is worth it.

\paragraph{The iterative route.} IRCOT* extracts knowledge triples from the retrieval results, asks what information is still missing, then it retrieves again, and repeats for up to three iterations (around 5–7 LLM calls, 3,500–5,500 tokens). It follows the IRCoT \cite{trivedi-etal-2023-interleaving} and is the most expensive one. The exact prompts are in the Appendix \ref{app:prompts}.

\paragraph{Three score predictors.} A yes/no classifier cannot express "Route B is slightly better than A, but Route C is even better," so with three routes, we change the target. We train three small \emph{score predictors}, one predictor for each route. Each one looks at the same six first-pass features as a 2-action RASER and answers a what-if question: if we sent this question to route r, what F1 would it get? We write these predictions:  "$\hat{f}_\stoproute(\mathbf{s})$, $\hat{f}_\pruneroute(\mathbf{s})$, and $\hat{f}_\iterroute(\mathbf{s})$. "The model class and training protocol are described together with the 2-action RASER in Section \ref{sec:setup}. 

\paragraph{Inference (cost-aware argmax).} Once we can estimate each route’s F1, it becomes easy to pick a route: for each route, take its predicted F1 and subtract a penalty for how many tokens it spends, then choose the route with the best score:

\begin{equation}
\begin{aligned}
r^\ast
&= \arg\max_{r \in \mathcal{R}}
\left[
\hat{f}_r(\mathbf{s}) - \lambda \overline{c}_r
\right], \\
\mathcal{R}
&= \{\stoproute, \pruneroute, \iterroute\}.
\end{aligned}
\label{eq:route_argmax}
\end{equation}


The $\lambda$ is an \emph{exchange rate} between tokens and accuracy about how much predicted \fone an extra token has to "buy" before the router is willing to spend it. At $\lambda=0$, tokens are free, so the router just takes the highest predicted \fone; as $\lambda$ grows, an expensive route must promise an even higher \fone to be chosen, and at a large enough $\lambda$ the router always stops. Turning this $\lambda$ from low to high traces out the cost-accuracy curve in Figure~\ref{fig:pareto_3route}: rather than providing a single fixed system, we give the operator choices and let them pick the token budget they are willing to pay.

On the same \emph{Young Man Luther} question used in Figure~\ref{fig:raser_mechanism}, the three predictors output
$\hat{f}_\stoproute=-0.04$, $\hat{f}_\pruneroute=0.33$, and
$\hat{f}_\iterroute=0.36$. \iterroute would win slightly on raw \fone. But \iterroute also costs about $500$ more tokens than \pruneroute, so after the cost penalty $\lambda\bar{c}_r$ (with $\lambda = 10^{-4}$ and route costs $\bar{c}_\stoproute = 1{,}222$, $\bar{c}_\pruneroute = 3{,}819$, $\bar{c}_\iterroute = 4{,}315$) the scores become $-0.16$, $-0.05$, and $-0.07$: \pruneroute wins and \methodthree picks the same route as \methodtwo. \iterroute's $0.03$ extra predicted \fone was not worth its extra tokens. ($\bar{c}_r$ and $\lambda$ are computed from training-fold data only; details in Section~\ref{sec:setup}.) An analysis varying the cost budget from $0.33$ to $1.00$ is in Appendix~\ref{app:sensitivity}.

A second example, where the IRCoT finds the results, is shown in Figure~\ref{fig:raser_ircot} (Appendix~\ref{app:workflow}). On the Achaemenid question (Llama-3.1-8B / MuSiQue), \methodtwo's escalation probability is $0.164$, just below the threshold $\theta=0.20$, so it keeps the one-shot \emph{"I don't know"}. \methodthree's regressors predict $\hat{f}_\iterroute=0.50$, much higher than \stoproute or \pruneroute, and after the cost penalty \iterroute still wins; two retrieve-extract rounds find the answer \emph{323 BC} (F1$=1$). This is the case where \methodtwo's threshold gates the question out and \methodthree's cost-aware argmax keeps \iterroute available. Prompts are in Appendix \ref{app:prompts}.

\methodtwo and \methodthree share the same features and the same
one-shot RAG results; they differ in classifier (single classifier vs.\ three regressors) and decision rule (threshold vs.\ cost-aware argmax). Both
are the main systems in this paper.

\section{Experimental Setup}
\label{sec:setup}

\paragraph{Benchmarks} We evaluate three multi-hop QA benchmarks with complementary properties. \textbf{MuSiQue} \citep{trivedi-etal-2022-musique} is our main benchmark because it was explicitly designed to reduce shortcut solutions and enforce connected reasoning. \textbf{2WikiMultiHopQA} \citep{ho-etal-2020-constructing} serves as a second benchmark with explicit reasoning paths and supporting evidence. \textbf{HotpotQA} \citep{yang-etal-2018-hotpotqa} is used as a control benchmark: many questions can already be handled well by one-shot retrieval, making it useful for testing whether a routing method remains restrained when bridge retrieval is less consistently necessary.

\paragraph{Retrieval} All methods use the same dense retriever:
Nomic-Embed-Text-v1.5 with cosine similarity \citep{nussbaum2025nomicembedtrainingreproducible}. Passages are sentence-segmented and chunked at 128 tokens with 16-token overlap. 

\paragraph{LLMs} We evaluate six LLMs: GPT-OSS-120B~\cite{openai2025gptoss120bgptoss20bmodel}, Mistral-Small-119B~\cite{mistral2026introducing-small4}, Gemma-3-31B-it~\cite{gemmateam2025gemma3technicalreport}, Llama-3.1-8B-Instruct, Llama-3-8B-Instruct~\cite{grattafiori2024llama3herdmodels}, and Phi-4-mini-instruct~\cite{microsoft2025phi4minitechnicalreportcompact}. All of them are queried through an OpenAI-compatible chat API at temperature 0.

\paragraph{GBM and hyperparameters} Both RASER routers are Gradient Boosting Machines (GBM) as implemented in scikit-learn
\citep{pedregosa2018scikitlearnmachinelearningpython}, with the same hyperparameters: 100 trees of max depth 3, learning rate 0.1, and subsample 0.8. RASER-2 is a single Gradient Boosting Classifier trained with log loss; RASER-3 is three Gradient Boosting Regressors, one for each
action, each trained with mean squared error (MSE).

\paragraph{Training label}
For \methodtwo, we label a question \textsc{Bridgeable} ($y=1$) only when the bridge route PRUNE improves the one-shot RAG answer by more than $\tau=0.1$ \fone:

\begin{equation}
y = \mathbb{I}\left[\fone_{\pruneroute} - \fone_{\stoproute} > \tau\right].
\label{eq:bridgeable_label}
\end{equation}

If \stoproute is already correct or no route helps, $y=0$; $y=1$ only when the PRUNE route
repairs a wrong first-RAG answer. For \methodthree, each regressor $\hat{f}_r$ is trained to predict the per-question \fone that route $r$ actually achieved on the training data. More details about how we trained RASER are provided in the appendix.

\section{Results}

Table~\ref{tab:main_results} summarizes the main result: fixed expensive retrieval policies often improve answer \fone, but they pay that cost on every question. RASER keeps high accuracy while avoiding many unnecessary, expensive LLM calls. We report the answer \fone and the average LLM tokens for each benchmark. The visualization of this table is Figure \ref{fig:visualization}, shown in the Appendix \ref{app: results_visualization}.

Fixed expensive baselines improve some questions, but they pay that cost for every question. Always running \pruneroute spends roughly $3\times$ the tokens of one-shot RAG, while its aggregate \fone gains are modest and uneven. \methodtwo targets exactly this regime: across six readers and three benchmarks, it stays within $0.026$ \fone of always-\pruneroute while using only $39$--$57\%$ of its tokens. This shows that a cheap recoverability decision can preserve most of the bridge benefit while removing much of the bridge cost. \methodthree adds an iterative route for settings where one-round bridging is not enough. Its value is LLM-dependent. Frontier LLMs \iterroute often add little beyond cheaper routes, and \methodthree stays close to \methodtwo. Mid-tier open LLMs, especially MuSiQue and 2Wiki, \iterroute has real marginal value; there \methodthree captures much of the always-\iterroute gain while spending far fewer tokens. Thus \methodthree should not be read as a universal replacement for \methodtwo, but as a higher-budget operating point when iterative retrieval is worth paying for.

\begin{figure}[t]
\centering
\includegraphics[width=0.5\textwidth]{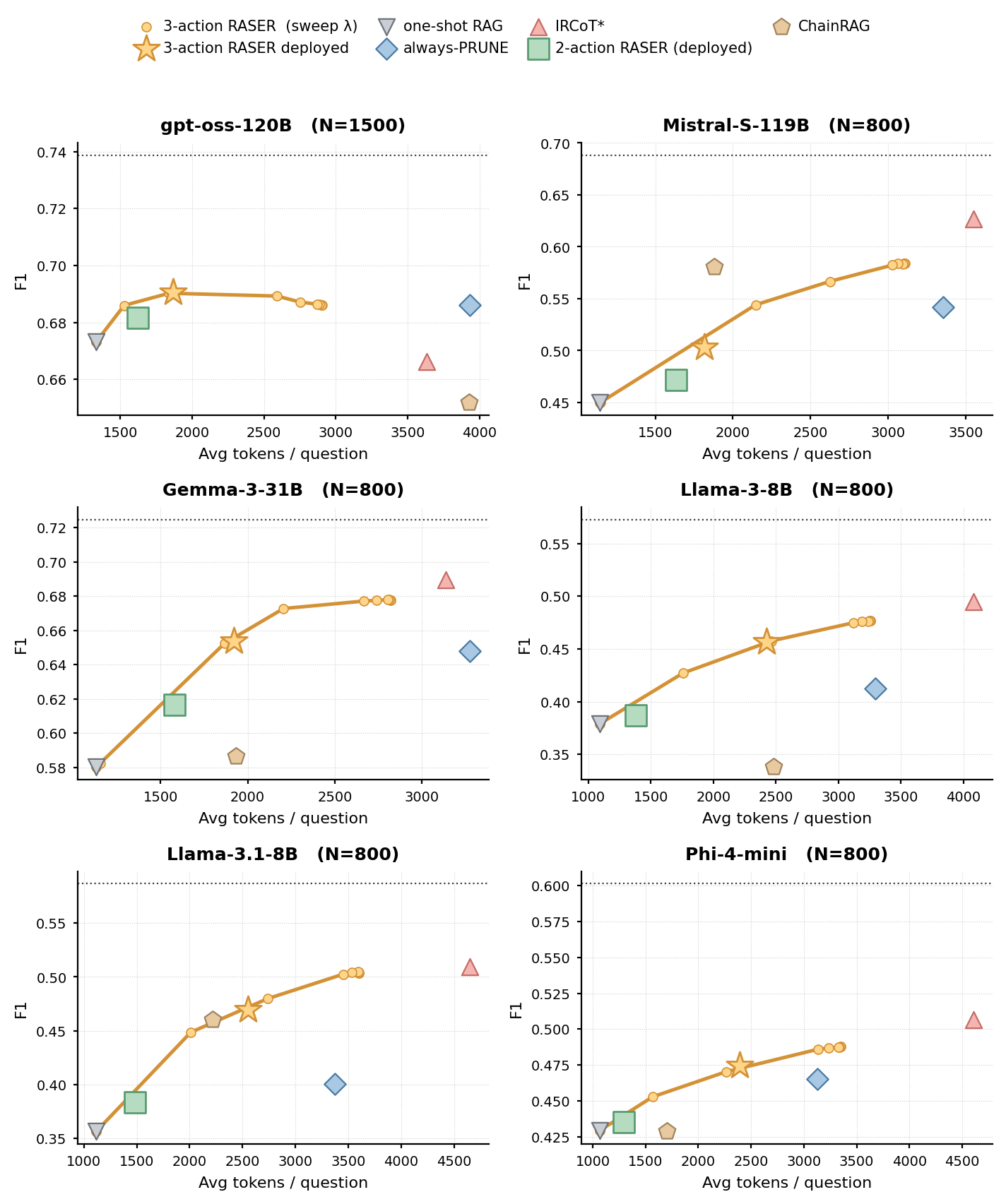}
\caption{Cost-\fone trade-off for each LLM across MuSiQue, 2WikiMultiHopQA, and HotpotQA (weighted by $N$ for each dataset). The \textbf{orange curve} is \methodthree as we sweep the cost penalty $\lambda$ from low to high; each point is one $\lambda$ setting. The \textbf{orange star} marks the deployed operating point (the \methodthree column of Table~\ref{tab:main_results}). Other methods include one-shot RAG, always-PRUNE, \iterroute, ChainRAG, and the deployed \methodtwo. The \emph{oracle}: the \fone a perfect router would be obtained if it could always pick the best of the three routes for each question. On small LLM Llama-3-8B and Phi-4-mini, \methodthree gets more \fone for the same tokens.}
\label{fig:pareto_3route}
\end{figure}

Finally, we check whether this conclusion depends on using a lightweight decomposition primitive. Replacing \decomproute with a controlled ChainRAG reimplementation under the same retriever, reader, and corpus changes GPT-OSS-120B \fone by at most $0.020$ per dataset at a similar token cost and does not change the method ranking. The result supports the same conclusion: RASER's gain comes from selective routing, not from inventing a stronger, more expensive route.

\begin{table*}[t]
\centering
\footnotesize
\setlength{\tabcolsep}{2.6pt}
\renewcommand{\arraystretch}{0.97}
\begin{tabular}{@{}l rr rr rr rr rr rr rr@{}}
\toprule
& \multicolumn{2}{c}{\stoproute}
& \multicolumn{2}{c}{a-\pruneroute}
& \multicolumn{2}{c}{\iterroute}
& \multicolumn{2}{c}{\decomproute}
& \multicolumn{2}{c}{ChainRAG$^\dagger$}
& \multicolumn{2}{c}{\textbf{\methodtwo}}
& \multicolumn{2}{c}{\methodthree}\\
\cmidrule(lr){2-3}\cmidrule(lr){4-5}\cmidrule(lr){6-7}\cmidrule(lr){8-9}\cmidrule(lr){10-11}\cmidrule(lr){12-13}\cmidrule(lr){14-15}
Reader / dataset & \fone & tk & \fone & tk & \fone & tk & \fone & tk
& \fone & tk & \fone & tk & \fone & tk\\
\midrule
\multicolumn{15}{l}{\emph{GPT-OSS-120B} {\footnotesize( $N{=}500$ per dataset)}}\\
\;MuSiQue & 0.488 & 1.5k & 0.528 & 4.8k & 0.455 & 4.3k & 0.482 & 4.6k & 0.465 & 4.6k & \textbf{0.502} & \textbf{2.1k} & \textbf{0.510} & \textbf{2.2k}\\
\;2Wiki & 0.754 & 1.1k & 0.761 & 3.3k & 0.781 & 3.0k & 0.743 & 3.5k & 0.763 & 3.6k & \textbf{0.763} & \textbf{1.3k} & \textbf{0.774} & \textbf{1.6k}\\
\;HotpotQA & 0.777 & 1.4k & 0.769 & 3.7k & 0.763 & 3.6k & 0.724 & 4.1k & 0.727 & 3.6k & \textbf{0.779} & \textbf{1.5k} & \textbf{0.787} & \textbf{1.8k}\\
\midrule
\multicolumn{15}{l}{\emph{Mistral-S-119B} {\footnotesize($N{=}300/300/200$)}}\\
\;MuSiQue & 0.283 & 1.2k & 0.397 & 3.8k & 0.499 & 4.3k & 0.380 & 3.3k & 0.464 & 2.3k & \textbf{0.306} & \textbf{1.9k} & \textbf{0.408} & \textbf{2.4k}\\
\;2Wiki & 0.450 & 970 & 0.570 & 2.9k & 0.658 & 2.9k & 0.562 & 2.8k & 0.643 & 1.7k & \textbf{0.485} & \textbf{1.5k} & \textbf{0.451} & \textbf{979}\\
\;HotpotQA & 0.699 & 1.3k & 0.716 & 3.4k & 0.773 & 3.4k & 0.638 & 3.3k & 0.662 & 1.6k & \textbf{0.702} & \textbf{1.4k} & \textbf{0.723} & \textbf{2.2k}\\
\midrule
\multicolumn{15}{l}{\emph{Gemma-3-31B} {\footnotesize($N{=}300/300/200$)}}\\
\;MuSiQue & 0.393 & 1.2k & 0.503 & 3.7k & 0.577 & 3.8k & 0.360 & 3.0k & 0.445 & 2.3k & \textbf{0.458} & \textbf{2.1k} & \textbf{0.502} & \textbf{2.4k}\\
\;2Wiki & 0.630 & 962 & 0.694 & 2.8k & 0.722 & 2.5k & 0.664 & 2.5k & 0.625 & 1.8k & \textbf{0.656} & \textbf{1.2k} & \textbf{0.700} & \textbf{1.6k}\\
\;HotpotQA & 0.787 & 1.3k & 0.797 & 3.4k & 0.810 & 3.0k & 0.734 & 2.8k & 0.740 & 1.6k & \textbf{0.796} & \textbf{1.4k} & \textbf{0.812} & \textbf{1.8k}\\
\midrule
\multicolumn{15}{l}{\emph{Llama-3-8B} {\footnotesize($N{=}300/300/200$)}}\\
\;MuSiQue & 0.223 & 1.2k & 0.256 & 3.7k & 0.354 & 4.7k & 0.292 & 3.2k & 0.209 & 2.8k & \textbf{0.229} & \textbf{1.6k} & \textbf{0.309} & \textbf{2.8k}\\
\;2Wiki & 0.367 & 932 & 0.406 & 2.8k & 0.507 & 3.5k & 0.460 & 2.2k & 0.377 & 2.1k & \textbf{0.380} & \textbf{1.2k} & \textbf{0.472} & \textbf{2.1k}\\
\;HotpotQA & 0.632 & 1.2k & 0.656 & 3.3k & 0.688 & 4.1k & 0.621 & 3.5k & 0.474 & 2.5k & \textbf{0.635} & \textbf{1.3k} & \textbf{0.655} & \textbf{2.2k}\\
\midrule
\multicolumn{15}{l}{\emph{Llama-3.1-8B} {\footnotesize($N{=}300/300/200$)}}\\
\;MuSiQue & 0.236 & 1.2k & 0.264 & 3.8k & 0.376 & 5.1k & 0.276 & 3.2k & 0.312 & 2.6k & \textbf{0.253} & \textbf{1.5k} & \textbf{0.330} & \textbf{3.0k}\\
\;2Wiki & 0.307 & 954 & 0.387 & 2.9k & 0.532 & 4.0k & 0.353 & 2.3k & 0.518 & 1.9k & \textbf{0.355} & \textbf{1.4k} & \textbf{0.487} & \textbf{2.2k}\\
\;HotpotQA & 0.612 & 1.3k & 0.626 & 3.4k & 0.676 & 4.8k & 0.625 & 3.6k & 0.596 & 2.2k & \textbf{0.624} & \textbf{1.5k} & \textbf{0.654} & \textbf{2.4k}\\
\midrule
\multicolumn{15}{l}{\emph{Phi-4-mini} {\footnotesize($N{=}300/300/200$)}}\\
\;MuSiQue & 0.331 & 1.1k & 0.353 & 3.6k & 0.372 & 5.4k & 0.407 & 3.0k & 0.293 & 1.9k & \textbf{0.335} & \textbf{1.4k} & \textbf{0.357} & \textbf{2.7k}\\
\;2Wiki & 0.391 & 904 & 0.459 & 2.6k & 0.532 & 3.8k & 0.457 & 2.6k & 0.475 & 1.6k & \textbf{0.402} & \textbf{1.1k} & \textbf{0.474} & \textbf{2.3k}\\
\;HotpotQA & 0.634 & 1.2k & 0.643 & 3.2k & 0.670 & 4.7k & 0.577 & 3.2k & 0.564 & 1.6k & \textbf{0.635} & \textbf{1.4k} & \textbf{0.652} & \textbf{2.2k}\\
\bottomrule
\end{tabular}
\caption{Main results: answer \fone and mean tokens per question (k\,$=10^3$) for six LLMs and three datasets under the same settings. Held-out $N$ is shown in each LLM: GPT-OSS-120B uses $N{=}500$ per dataset; the other readers use $N{=}300/300/200$ for MuSiQue/2Wiki/HotpotQA. The routers this paper contributes to; \methodtwo keeps most of the bridge accuracy at a much lower cost, and \methodthree adds the iterative route, which has real improvements.}
\label{tab:main_results}
\end{table*}

\subsection{Why does iterative retrieval help small-size LLMs more?}
\label{sec:memory_check}

The LLMs comparison in Table~\ref{tab:main_results} shows that larger-sized LLMs gain little from \iterroute, while middle- and small-sized LLMs benefit from it. One explanation is that larger-sized LLMs rely less on retrieved evidence. They may answer some questions from their memory even when the retrieved passage changes. Because these LLMs are trained with that information.

We test this on 200 MuSiQue questions. For each question, we edit the retrieved passages that contain the answer, we replace the correct answer with a plausible incorrect one, and then we run \stoproute again. If the LLM outputs the new, incorrect answer, it follows the retrieved evidence.
If it keeps the original gold answer, it is likely relying on internal memory.




\begin{table}[t]
\centering
\scriptsize
\setlength{\tabcolsep}{3.0pt}
\renewcommand{\arraystretch}{1.05}

\begin{tabularx}{\linewidth}{@{}lYYYY@{}}
\toprule
LLM & Edited & Gold & Abstain & Other \\
\midrule
GPT-OSS-120B & 41\% & \textbf{38\%} & 12\% & 9\% \\
Gemma-3-31B  & 52\% & 21\%          & 18\% & 9\% \\
Llama-3-8B   & 67\% & 9\%           & 12\% & 12\% \\
\bottomrule
\end{tabularx}

\caption{Memory check on 200 MuSiQue questions. \textit{Edited} means the reader follows the corrupted retrieved passage and outputs the edited wrong answer. \textit{Gold} means the reader keeps the original gold answer instead of edited evidence, suggesting it relies on internal memory rather than retrieved evidence.}
\label{tab:memory_check}
\end{table}

The results show GPT-OSS-120B keeps the original answer on $38\%$ of edited questions, while Llama-3-8B only $9\%$. This explains why larger LLMs gain less from iterative retrieval; many of their one-shot RAG successes are not true retrieval successes, so extra retrieval has less space to help. Smaller LLMs follow the retrieved text more often, making retrieval depth more important.

\subsection{Features Ablation}
\label{sec:feature_ablation}

Which input features mainly drive the RASER Table~\ref{tab:feature_importance} shows that tree features are important in all LLMs. Higher importance means the router uses that feature more often when making routing decisions. The \texttt{score\_gap} and \texttt{score\_top1}, account for $0.73$ of the total importance, showing that the router primarily relies on whether the first retrieval step produced clear and reliable evidence.
One-shot RAG answer features are secondary: \texttt{confidence} and \texttt{ans\_len} together contribute $0.24$.
By contrast, \texttt{bridge\_cues} and \texttt{qtype} contribute only $0.03$, suggesting that RASER learns to escalate when the retrieved evidence and draft answer jointly indicate that the one-shot RAG is unreliable.

\begin{table}[t]
\centering
\footnotesize
\setlength{\tabcolsep}{4.2pt}
\renewcommand{\arraystretch}{0.92}
\begin{tabularx}{\linewidth}{@{}l r >{\raggedright\arraybackslash}X@{}}
\toprule
Feature & Importance & Plain meaning \\
\midrule
\texttt{score\_gap}    & 0.38 & top-1 vs. top-5 score gap \\
\texttt{score\_top1}   & 0.35 & top-1 retrieved passage's score \\
\texttt{confidence}    & 0.12 & rule-based confidence \\
\texttt{ans\_len}      & 0.12 & lengthen of the draft answer \\
\texttt{bridge\_cues}  & 0.02 & bridge-like words in the question \\
\texttt{qtype}         & 0.01 & question type \\

\bottomrule
\end{tabularx}
\caption{Feature importance for RASER.}
\label{tab:feature_importance}
\end{table}

\subsection{Classifiers Ablation}

We also tried other classifiers. We compare the two-route GBM with XGBoost, LightGBM, CatBoost, scaled logistic regression, and a small MLP. RASER might also work in other classifiers. We applied Gradient Boosting Machine because it is the simplest and most widely available default. Some results are shown in Table \ref{tab:classifier_ablation}. Full classifier comparisons are reported in Appendix~\ref{app:classifier_ablation}.

\begin{center}
\centering
\scriptsize
\setlength{\tabcolsep}{4.2pt}
\renewcommand{\arraystretch}{0.82}
\begin{tabular}{@{}llccc@{}}
\toprule
Router & Model & \fone & Tokens & Route \\
\midrule
\methodtwo & GBM    & 0.520 & 1514 & 86/14 \\
           & LogReg & 0.531 & 1549 & 85/15 \\
           & MLP    & 0.530 & 1597 & 82/18 \\
\midrule
\methodthree & GBM     & 0.562 & 2157 & 64/12/24 \\
             & XGBoost & 0.562 & 2094 & 67/10/23 \\
             & Ridge   & 0.546 & 1844 & 76/5/18 \\
\bottomrule
\end{tabular}
\captionof{table}{Part of Classifier ablation. GBM is not the best, but it is within the same \fone range as the alternatives while being simple and reproducible.}
\label{tab:classifier_ablation}
\end{center}


\section{Discussion}
\label{sec:discussion}

\paragraph{When Should We Use RASER-3?}

3-action RASER is the cost-aware router. Its purpose is not to replace 2-action but to choose among one-shot RAG, PRUNE, and IRCOT* when the application is willing to spend more for
accuracy. This matters when iterative retrieval is useful. On the middle-sized LLMs, IRCOT* improves F1 over one-shot RAG and bridging retrieval PRUNE (for example, Llama-3.1-8B on MuSiQue:
IRCOT* 0.376 vs. PRUNE 0.264). In this case, 3-action RASER reaches 87–97\% of always-IRCOT* F1 while spending only 47–62\% of its tokens.

The RASER-3 results in Table \ref{tab:main_results} are the results of the specific $\lambda$ we deployed. not RASER-3 itself. In our setup $\lambda$ is the cost penalty and it is in the cost-aware argmax (Eq.~\ref{eq:route_argmax}); Figure \ref{fig:pareto_3route} shows turning $\lambda$ higher also applies more tokens. Small $\lambda$ has higher \fone with higher token cost; large $\lambda$ performs worse than \methodtwo. \methodthree is not a fixed system at all, but it is a cost dial. It behaves based on your budget tokens. Appendix~\ref{app:router_diag} checks both routers. \methodtwo escalates the questions where \pruneroute actually helps. \methodthree uses \iterroute more often for the LLMs and datasets where \iterroute actually gives an \fone gain.

\paragraph{Comparison to other Baselines}

RASER is lightweight by design. The router reads six features already available after the one-shot RAG pass and makes no additional LLM calls before deciding whether to escalate. It also does not require pre-built knowledge graphs, pre-trained retrievers, or an LLM judge for routing. This distinguishes RASER from heavier iterative-retrieval systems. In our experiments, \iterroute and \decomproute are controlled methods based on literature, not full reproductions of KiRAG or ChainRAG. The contribution of RASER is not a new, expensive retrieval method; it is a recoverability-aware decision layer that decides when bridge retrieval is worthy.

\section{Conclusion}

Multi-hop QA is usually treated as a "retrieve more" problem, but our analysis shows that only a small portion of questions benefit from extra retrieval. Then \method asks the research question, "When is extra retrieval worth it?" It is a small classifier that runs after one-shot RAG, reads six features, and decides whether to escalate to a bridge step (\methodtwo) or to iterative retrieval (\methodthree), with no extra LLM calls. Across six LLMs and three benchmarks, \methodtwo and \methodthree reach \fone comparable to and sometimes higher than the strongest baselines, while spending far fewer tokens. It shows deciding \emph{when} to retrieve more is cheaper and more useful than always retrieving more.

\section*{Limitations}
\label{sec:limits}

\paragraph{Baselines are simplified versions} We did not completely run the original KiRAG or ChainRAG, but a simplified version of them. Our \iterroute, \decomproute, and ChainRAG baselines share the same retriever and LLM as RASER, so the comparison is fair in our setup. But the numbers should not be compared directly to the original papers' results. KiRAG, for example, has trained some heavy reasoning components that we did not reproduce.

\paragraph{Small evaluation size and reader memory} We evaluate $200$--$500$ questions for each LLM and dataset, so the per-cell \fone numbers have some noise. The memory check in \S\ref{sec:memory_check} also shows that large-sized LLMs sometimes answer from memory instead of the retrieved passage. So \fone on those LLMs reflect both retrieval and the LLM's own knowledge, not retrieval alone. But RASER performs well on smaller-sized LLMs, which have less memory.

\bibliography{custom}

\appendix


\section{Supplemental Material Statement}

The anonymized code and scripts needed to reproduce the experiments are available at:\url{https://github.com/YuyangLi99/RASER_2026}.
The repository contains the implementation of the RASER pipeline, retrieval methods and routing systems, prompts, and configuration files, which are used to reproduce the main experiments and RASER system.

\section{RASER workflow}
\label{app:workflow}

\begin{figure*}[t]
\centering
\includegraphics[width=0.95\textwidth]{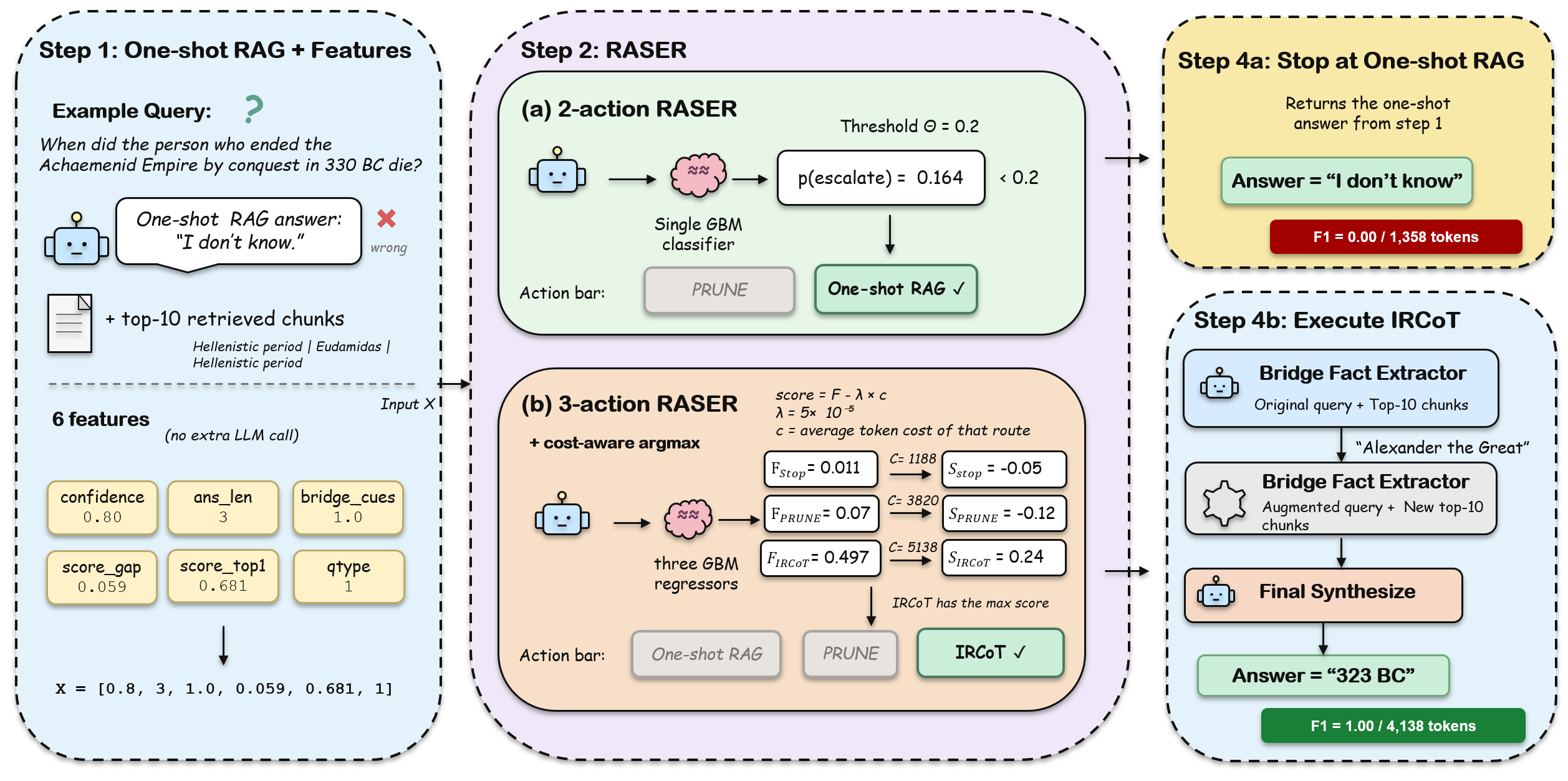}
\caption{\textbf{\method on a question where the two routers disagree.} Same pipeline as Figure~\ref{fig:raser_mechanism}, on the Achaemenid question (Llama-3.1-8B / MuSiQue). \methodtwo's escalation probability $0.164$ is below the threshold $\theta = 0.20$, so it returns the one-shot \emph{"I don't know"} (Step~4a, F1$=0$). \methodthree picks \iterroute via the cost-aware argmax ($\hat{f}_\iterroute$ stays highest even after the cost penalty), and two retrieve-extract rounds with the Bridge Fact Extractor find \emph{323 BC} (Step~4b, F1$=1$). Full step-by-step trace in Example~3 below.}
\label{fig:raser_ircot}
\end{figure*}

We show \method's full pipeline on three example questions on MuSiQue with Llama-3.1-8B, one question for each route: one example is about the one-shot RAG answer being already correct, one example is about extra bridge steps fixing the answer (\pruneroute), and the final example is about only iterative retrieval being able to find the missing fact and generate the correct answers(\iterroute). Figure \ref{fig:raser_ircot} above visualizes the third example (the IRCoT case); the other two are discussed in the text only. All F1 and token costs come directly after the runs.

\subsection{Workflow prompts}
\label{app:prompts}

The prompts below are what we actually send to the LLM. 

\paragraph{Step 1—One-shot RAG.} After running the first time RAG, we get the top 10 retrieved texts from the dense retriever; then, we transfer this information as the following example prompts to the LLM:

\promptbox{Prompt: One-shot RAG}{%
Answer the following question concisely based on the given context.
If you cannot find the answer, reply ``I don't know''.\\[3pt]
\textbf{Context}: \emph{\{retrieved chunks\}}\\
\textbf{Question}: \emph{\{q\}}\\
\textbf{Answer (be concise, just the answer)}:
}

\paragraph{\pruneroute -- Bridge Proposer.} After RASER, we ask LLM to extract up to two bridge candidates as JSON; the prompts are shown as follows:

\promptbox{Prompt: Bridge Proposer}{%
You are decomposing a multi-hop question to find intermediate bridge
entities. Identify up to 2 intermediate facts that could help answer
the question.\\[3pt]
For each, output a JSON object on its own line with these fields:
\begin{itemize}\setlength\itemsep{0pt}\setlength\parskip{0pt}
\item \texttt{bridge\_entity}: the intermediate fact ($\le$ 5 words)
\item \texttt{bridge\_relation}: what role it plays
\item \texttt{missing\_slot}: what is still missing after this bridge
\item \texttt{confidence}: $0.0$--$1.0$
\end{itemize}
\textbf{Passages}: \emph{\{retrieved chunks\}}\\
\textbf{Question}: \emph{\{q\}}\\
\textbf{Current answer}: \emph{\{one-shot answer\}}\\
\textbf{Bridge proposals (one JSON per line)}:
}

\paragraph{\pruneroute -- Branch Verifier.} Rule-based verifier, no
LLM call. For each bridge, compute (i) \emph{novelty} (fraction of new
chunks vs.\ one-shot), (ii) \emph{support} (fraction of new chunks
mentioning the bridge), (iii) \emph{info\_gain} (1 -- Jaccard overlap).
A branch is dropped if novelty $<\!0.05$ or support $<\!0.05$.

\paragraph{\pruneroute -- Final Synthesizer.} We merge the
surviving chunks with the one-shot chunks and tell the LLM to bridge the fact that it can trust:

\promptbox{Prompt: Final Synthesizer}{%
Answer the following question concisely based on the given context.
If you cannot find the answer, reply ``I don't know''.\\[3pt]
\textbf{Intermediate facts already established by decomposition
(treat as authoritative bridge entities)}:\\
\emph{\{- bridge\_1\}\\- \{bridge\_2\}}\\[3pt]
\textbf{Context}: \emph{\{merged branch + one-shot chunks\}}\\
\textbf{Question}: \emph{\{q\}}\\
\textbf{Answer (be concise, just the answer)}:
}

\paragraph{\iterroute -- iterative retrieval.} Up to three rounds:
extract a short fact from the current chunks, append it to the query,
re-retrieve. The prompt shows below:

\promptbox{Prompt: \iterroute fact extractor}{%
You are decomposing a multi-hop question. Read the passages and
Extract one intermediate fact that helps answer the question. The
fact MUST be a short answer string (entity name, date, place ---
at most 5 words). NOT a sentence. NOT an explanation. NO
reasoning.\\[3pt]
\textbf{Examples of valid replies}: \emph{Tracy McConnell / 1973 /
Mississippi River / DONE}\\
\textbf{Examples of INVALID replies}: \emph{``The answer is X
because\dots''} / multi-line essays.\\[3pt]
\textbf{Passages}: \emph{\{retrieved chunks\}}\\
\textbf{Facts already found}: \emph{\{prior bridges\}}\\
\textbf{Question}: \emph{\{q\}}\\
\textbf{Fact}:
}

\noindent The iteration stops when the LLM says \texttt{DONE}, after
3 rounds, or when new chunks overlap the previous round's chunks by Jaccard
$>\!0.6$. The final answer is then synthesized with the same Final
Synthesizer prompt as \pruneroute.

\subsection{Example 1: one-shot RAG already works (Llama-3.1-8B / MuSiQue)}

Many multi-hop questions are actually solved by one-shot RAG, because
the top retrieved chunk happens to contain both hops. The question
below has a two-hop structure---which state hosts the building, then
who that state voted for in 2016---but the top chunk is the Wikipedia
page for the building (which names New Hampshire), and the next chunk
is the 2016 election page. The LLM reads off Hillary Clinton from
these two chunks. Both routers see a high \texttt{score\_top1} and
don't escalate. Running \pruneroute or \iterroute would give the same
answer at $3$--$4\times$ the token cost.

\noindent
\fcolorbox{stopframe}{stopfill}{%
\begin{minipage}{0.95\linewidth}\small
\textbf{Question.} Who did the state where Christian Science Pleasant
View Home was located vote for in 2016?\par
\textbf{Gold.} \emph{Hillary Clinton.}
\par\smallskip\rule{\linewidth}{0.3pt}\par\smallskip

\textbf{Step 1 -- One-shot RAG.}\hfill\colorbox{goodgreen}{F1 $=$ 1.00}~~\textbf{1{,}231 tokens}\par
\smallskip
Top-3 retrieved chunks: \emph{Christian Science Pleasant View Home}
(the building, Concord NH); \emph{2016 U.S.\ presidential election};
\emph{2016 U.S.\ presidential election in Texas}. The LLM reads off
\emph{Hillary Clinton}.
\par\smallskip\rule{\linewidth}{0.3pt}\par\smallskip

\textbf{Step 2 -- Features.}\par
\texttt{confidence}\,=\,0.6,\
\texttt{ans\_len}\,=\,2,\
\texttt{bridge\_cues}\,=\,1,\
\texttt{score\_gap}\,=\,0.149,\
\texttt{score\_top1}\,=\,0.778,\
\texttt{qtype}\,=\,entity.\par
\smallskip
$\mathbf{x} = [\,0.6,\ 2,\ 1.0,\ 0.149,\ 0.778,\ 0\,]$
\par\smallskip\rule{\linewidth}{0.3pt}\par\smallskip

\textbf{Step 3 -- Routers.}\par\smallskip
\methodtwo: \quad $p(\textsc{Escalate}) = 0.010 \;<\; \theta = 0.20$ \quad$\Rightarrow$\quad \colorbox{stopfill}{\strut~\stoproute~$\checkmark$~}\par\smallskip
\methodthree: \quad
$\hat{f}_\stoproute = 0.378$,\;
$\hat{f}_\pruneroute = 0.173$,\;
$\hat{f}_\iterroute = 0.244$.\par
After cost penalty ($\lambda = 5{\times}10^{-5}$): $0.32$,\,$-0.02$,\,$-0.01$ \quad$\Rightarrow$\quad \colorbox{stopfill}{\strut~\stoproute~$\checkmark$~}
\par\smallskip\rule{\linewidth}{0.3pt}\par\smallskip

\textbf{Final.} Both routers choose one-shot RAG. They save $2{,}727$ tokens per
question vs.\ always-\pruneroute and $3{,}991$ vs.\ always-\iterroute with finding the correct answer.
\end{minipage}}

\subsection{Example 2: bridge route PRUNE fixes the answer (Gemma-3-31B / MuSiQue)}

Some multi-hop questions need one bridge step to find the answer. The question below asks which province borders the province that hosts Lago District. The top-retrieved chunk is the Lago District page (it says Lago District is in Niassa Province), but no chunk specifies what borders Niassa. So, one-shot RAG answers, "I don't know." Both routers escalate to \pruneroute: \methodtwo's escalation probability is $0.83$, well above the $0.20$ threshold; \methodthree's regressor also predicts \pruneroute will get the highest F1. The Bridge Proposer extracts \emph{Niassa Province} as the bridge, the Branch Retriever pulls in Niassa-related chunks, and the Final Synthesizer reads off \emph{Cabo Delgado Province}.

\noindent
\fcolorbox{pruneframe}{prunefill}{%
\begin{minipage}{0.95\linewidth}\small
\textbf{Question.} What province shares a border with the province where Lago District is located?\par
\textbf{Gold.} \emph{Cabo Delgado Province.}
\par\smallskip\rule{\linewidth}{0.3pt}\par\smallskip

\textbf{Step 1 -- One-shot RAG.}\hfill\colorbox{badred}{F1 $=$ 0.00}~~\textbf{1{,}124 tokens}\par
\smallskip
Top-3 retrieved chunks: \emph{Lago District} (says it is in Niassa
Province), \emph{Lago Verde (Queyras)}, \emph{Gallaratese}. No chunk
says which province borders Niassa. LLM answers \emph{``I don't
know''}.
\par\smallskip\rule{\linewidth}{0.3pt}\par\smallskip

\textbf{Step 2 -- Features.}\par
\texttt{confidence}\,=\,0.0,\
\texttt{ans\_len}\,=\,3,\
\texttt{bridge\_cues}\,=\,0,\
\texttt{score\_gap}\,=\,0.268,\
\texttt{score\_top1}\,=\,0.837,\
\texttt{qtype}\,=\,entity.\par
\smallskip
$\mathbf{x} = [\,0.0,\ 3,\ 0.0,\ 0.268,\ 0.837,\ 0\,]$
\par\smallskip\rule{\linewidth}{0.3pt}\par\smallskip

\textbf{Step 3 -- Routers.}\par\smallskip
\methodtwo: \quad $p(\textsc{Escalate}) = 0.830 \;\ge\; \theta = 0.20$ \quad$\Rightarrow$\quad \colorbox{prunefill}{\strut~\pruneroute~$\checkmark$~}\par\smallskip
\methodthree: \quad
$\hat{f}_\stoproute = 0.506$,\;
$\hat{f}_\pruneroute = 0.888$,\;
$\hat{f}_\iterroute = 0.807$.\par
After cost penalty ($\lambda = 1{\times}10^{-4}$): $0.39$,\,$0.52$,\,$0.42$ \quad$\Rightarrow$\quad \colorbox{prunefill}{\strut~\pruneroute~$\checkmark$~}
\par\smallskip\rule{\linewidth}{0.3pt}\par\smallskip

\textbf{Step 4 -- \pruneroute execution.}\hfill\colorbox{goodgreen}{F1 $=$ 1.00}~~\textbf{3{,}603 tokens}\par
\smallskip
Bridge Proposer extracts \emph{Niassa Province} (the province
containing Lago District). Branch Retriever pulls Niassa-related
chunks. Branch Verifier keeps the branch (support $=$ 0.78). Final
Synthesizer combines the bridge with the chunks and answers
\emph{Cabo Delgado Province}.
\end{minipage}}

\subsection{Example 3: iterative retrieval IRCOT* is needed (Llama-3.1-8B / MuSiQue)}

Some questions need more than one bridge step. The question below
asks for the death date of \emph{"the person who ended the Achaemenid
Empire by conquest in 330 BC"}. The bridge is Alexander the Great,
but identifying him is not enough --- we still need to retrieve his
biography to find when he died. This is exactly where the two routers
diverge. \methodtwo's escalation probability is $0.164$, just below
the $0.20$ threshold, so it stays with the one-shot \emph{"I don't
know"} (F1 $=$ 0). \methodthree is not gated by a threshold; its
three regressors predict $\hat{f}_\iterroute = 0.50$, much higher than
$\hat{f}_\stoproute = 0.01$ and $\hat{f}_\pruneroute = 0.07$, and after the cost penalty \iterroute still wins. \iterroute then iterates:
round 1 extracts Alexander the Great, round 2 retrieves his
biography and reads off \emph{323 BC} (F1 $=$ 1.00). This is the
clearest case in our held-out set where the threshold hurts \methodtwo
and the cost-aware argmax saves \methodthree.

\noindent
\fcolorbox{iterframe}{iterfill}{%
\begin{minipage}{0.95\linewidth}\small
\textbf{Question.} When did the person who ended the Achaemenid
Empire by conquest in 330 BC die?\par
\textbf{Gold.} \emph{323 BC} (Alexander the Great).
\par\smallskip\rule{\linewidth}{0.3pt}\par\smallskip

\textbf{Step 1 -- One-shot RAG.}\hfill\colorbox{badred}{F1 $=$ 0.00}~~\textbf{1{,}358 tokens}\par
\smallskip
Top-3 retrieved chunks: \emph{Hellenistic period}, \emph{Eudamidas I},
\emph{Hellenistic period} (second chunk) --- related to the era but
never name Alexander or his death date. LLM answers
\emph{``I don't know''}.
\par\smallskip\rule{\linewidth}{0.3pt}\par\smallskip

\textbf{Step 2 -- Features.}\par
\texttt{confidence}\,=\,0.8 (date prior),\
\texttt{ans\_len}\,=\,3,\
\texttt{bridge\_cues}\,=\,1,\
\texttt{score\_gap}\,=\,0.059,\
\texttt{score\_top1}\,=\,0.681,\
\texttt{qtype}\,=\,date.\par
\smallskip
$\mathbf{x} = [\,0.8,\ 3,\ 1.0,\ 0.059,\ 0.681,\ 1\,]$
\par\smallskip\rule{\linewidth}{0.3pt}\par\smallskip

\textbf{Step 3 -- Routers (here they disagree).}\par\smallskip
\methodtwo: \quad $p(\textsc{Escalate}) = 0.164 \;<\; \theta = 0.20$ \quad$\Rightarrow$\quad \colorbox{stopfill}{\strut~\stoproute~}\par\smallskip
\methodthree: \quad
$\hat{f}_\stoproute = 0.011$,\;
$\hat{f}_\pruneroute = 0.070$,\;
$\hat{f}_\iterroute = 0.497$.\par
After cost penalty ($\lambda = 5{\times}10^{-5}$): $-0.05$,\,$-0.12$,\,$0.24$ \quad$\Rightarrow$\quad \colorbox{iterfill}{\strut~\iterroute~$\checkmark$~}
\par\smallskip\rule{\linewidth}{0.3pt}\par\smallskip

\textbf{Step 4a -- what \methodtwo runs.}\hfill\colorbox{badred}{F1 $=$ 0.00}~~\textbf{1{,}358 tokens}\par
\smallskip
\methodtwo keeps the one-shot \emph{``I don't know''}. The probability
$0.164$ is just below $0.20$, so the router does not escalate. F1 $=$ 0.
\par\smallskip\rule{\linewidth}{0.3pt}\par\smallskip

\textbf{Step 4b -- what \methodthree runs.}\hfill\colorbox{goodgreen}{F1 $=$ 1.00}~~\textbf{4{,}138 tokens}\par
\smallskip
\methodthree runs \iterroute. Round 1 retrieves \emph{Hellenistic
period} chunks and extracts the fact \emph{``Alexander the Great''};
round 2 re-retrieves with that fact added to the query, hitting
Alexander's biography, and extracts \emph{``323 BC''}. Final answer:
\emph{323 BC}.
\end{minipage}}

\noindent On this question \iterroute is both cheaper \emph{and} more
accurate than \pruneroute would have been ($4{,}138$ vs.\ $4{,}306$
tokens; F1 $1.00$ vs.\ $0.00$), because the second hop's evidence
needs a query rewrite that \pruneroute's single re-retrieval cannot do.

\subsection{Take-away}

\method can do three things with its budget: spend no extra LLM calls (Example~1, \stoproute, 1.2k tokens), spend bridge retrieval calls (Example~2, \pruneroute, 2.2k tokens), or run iterative retrieval IRCoT (Example~3, \iterroute, 4.1k tokens). \methodtwo picks between the first two actions. \methodthree picks among all three actions. The router's job is to spend the smallest budget that still gets the correct question.

\section{Baseline Workflows}
\label{app:baseline_workflows}

\begin{figure*}[t]
\centering
\includegraphics[width=0.8\textwidth]{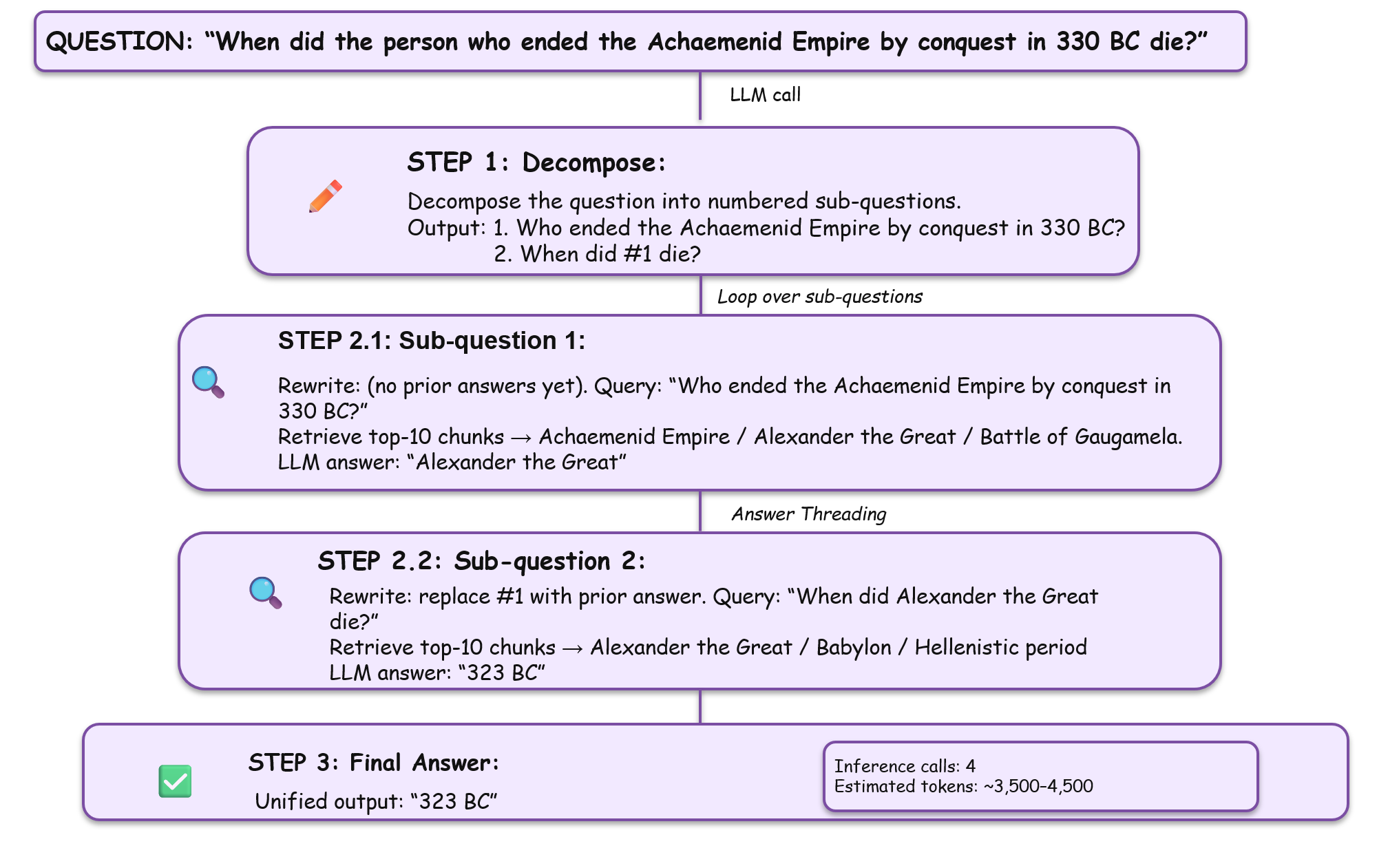}
\caption{\textbf{\decomproute (Self-Ask) baseline on the Achaemenid question.} Step 1: one LLM call decomposes the question into two sub-questions. Step 2.1: Retrieve and answer the first sub-question (\emph{Alexander the Great}). Step 2.2: Rewrite the second sub-question by substituting \texttt{\#1} with the first answer (\emph{answer threading}), retrieve and answer (\emph{323 BC}). Step 3: The last sub-answer is the final output. }
\label{fig:selfask}
\end{figure*}

\begin{figure*}[t]
\centering
\includegraphics[width=0.8\textwidth]{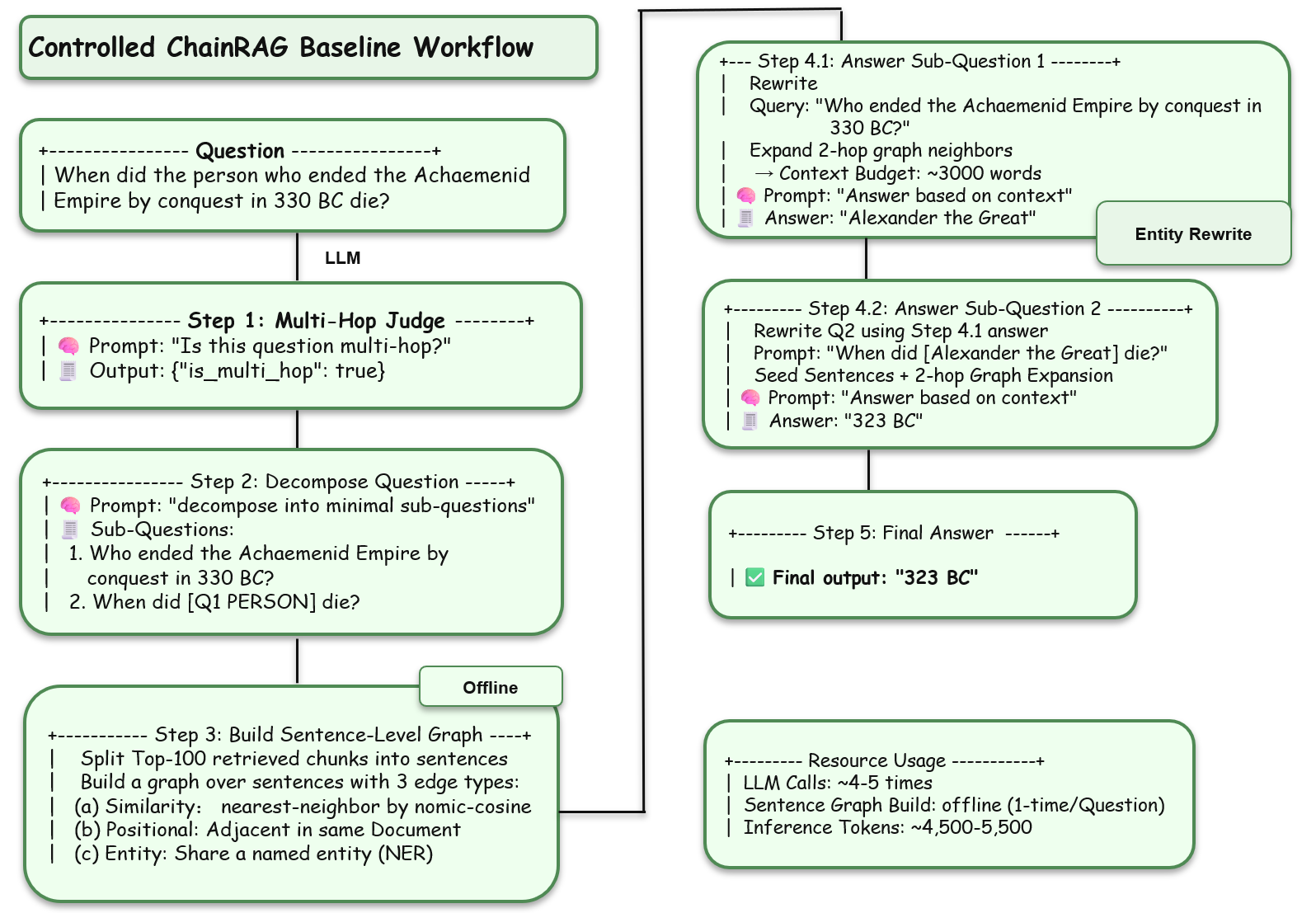}
\caption{\textbf{ChainRAG baseline on the Achaemenid question.} The pipeline adds a multi-hop judge (Step~1) and an offline sentence-level graph build (Step~3) with three edge types (similarity, positional, entity) on top of decomposition. Each sub-question retrieves seed sentences and expands their 2-hop neighbors on the graph. Sub-question 2 is rewritten by replacing the reference to the first sub-answer.}
\label{fig:chainrag}
\end{figure*}

For comparison with the \method workflow in Figure~\ref{fig:raser_ircot}, Figures~\ref{fig:selfask} and~\ref{fig:chainrag} show how the two LLM-driven baselines from Table~\ref{tab:main_results}, \decomproute (Self-Ask-style) and ChainRAG, process the same Achaemenid question.

\paragraph{\decomproute (Self-Ask).} One LLM call decomposes the question into numbered sub-questions, with later sub-questions referencing earlier answers via \texttt{\#N}. For each sub-question in order, we rewrite it with prior answers substituted in, retrieve the top-10 chunks, and ask the LLM. The last sub-answer is the final output. On the Achaemenid question, this takes 4 LLM calls and about $3{,}500$--$4{,}500$ tokens.

\paragraph{ChainRAG.} ChainRAG adds two steps on top of Self-Ask. First, a multi-hop judge checks whether the question is multi-hop; only multi-hop questions go through decomposition. Second, before processing the sub-questions, it builds a sentence-level graph over the top-100 retrieved chunks. The graph has three edge types: similarity (k-nearest neighbors by nomic cosine), positional (adjacent in the same document), and entity (sharing a named entity). Each sub-question is answered by retrieving seed sentences and expanding their 2-hop neighbors on the graph. Later sub-questions are rewritten with prior answers (\emph{entity completion}). On the Achaemenid question, this takes 4--5 LLM calls and about $4{,}500$--$5{,}500$ tokens.

\section{Main results visualization}
\label{app: results_visualization}

\begin{figure}[t]
\centering
\includegraphics[width=0.5\textwidth]{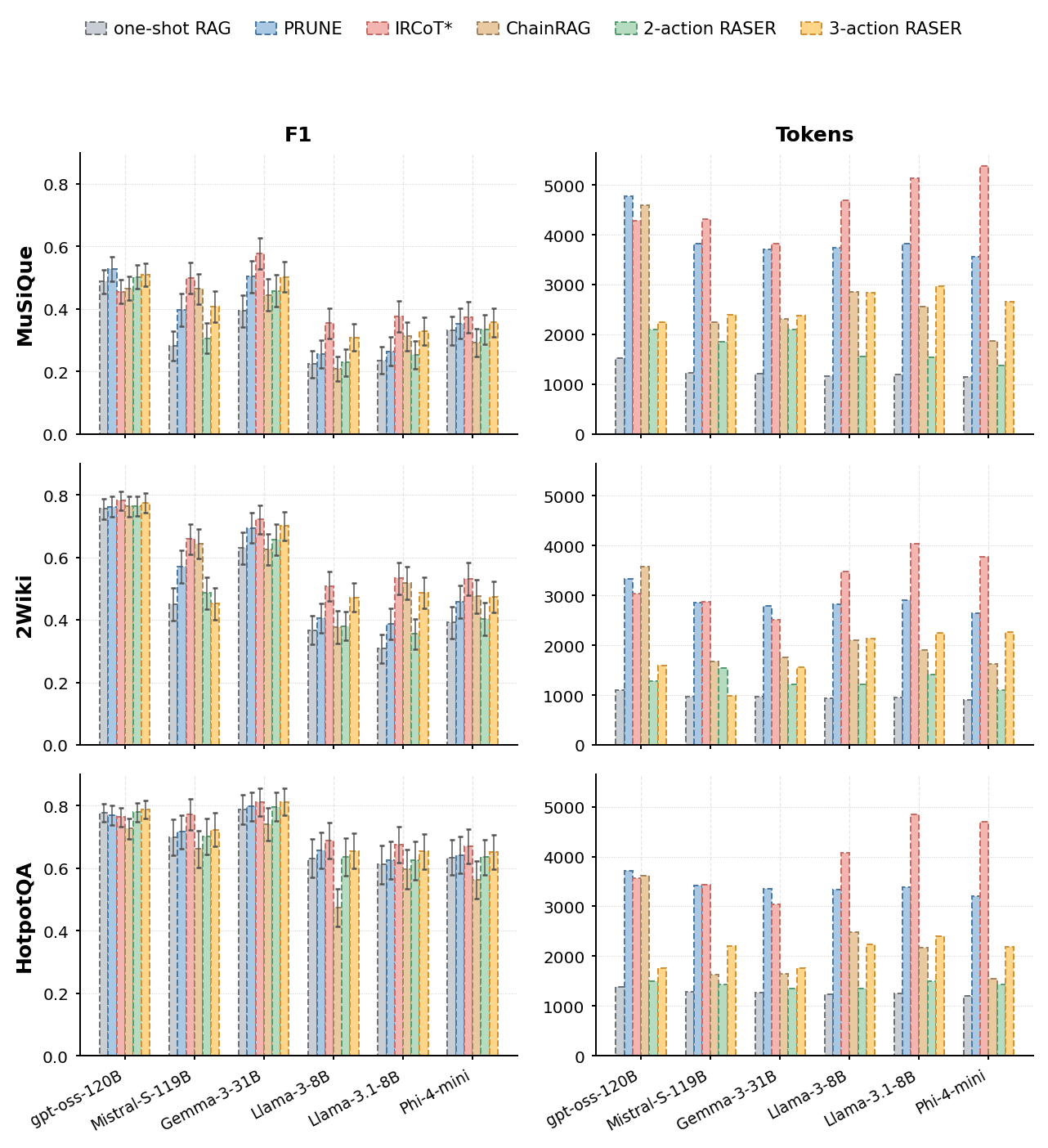}
\caption{Visualization of Table~\ref{tab:main_results}, \textbf{Bottom:} average tokens for each benchmark. The Self-Ask\textsuperscript{*} baseline is in the table \ref{tab:main_results}.}
\label{fig:visualization}
\end{figure}

Figure \ref{fig:visualization} shows the visualization of Table \ref{tab:main_results}. Which provides a clear vision about how RASER performs compared with other baselines.

\section{Router Diagnose}
\label{app:router_diag}

\paragraph{\methodtwo.}
Only $9$--$16\%$ of questions actually benefit from \pruneroute over
\stoproute. \methodtwo escalates $14\%$ of the time, close to that rate
---it neither over- nor under-escalates. Among questions the router
assigns high probability, the bridgeable rate is about $3\times$ higher
than among low-probability questions, so the score is meaningful.

\paragraph{\methodthree.}
Pooled across LLMs and datasets, \methodthree sends $64\%$ of questions
to \stoproute, $12\%$ to \pruneroute, and $24\%$ to \iterroute. The mix
shifts per cell: cells where \iterroute is the best fixed route
(mid-tier LLMs on MuSiQue/2Wiki) get $>30\%$ \iterroute; cells where
\stoproute is already strong (frontier LLMs on HotpotQA) stay at
$>80\%$ \stoproute.

\section{Classifier Ablation}
\label{app:classifier_ablation}

Table~\ref{tab:model_ablation} is the full version of the classifier
ablation we summarized in the main text (Table~\ref{tab:classifier_ablation}). The main text shows only 3 of
the 6 classifiers per router for space; here, we include all of them. The setup is identical: the same six features, the same LLM and dataset, the same threshold $\theta = 0.20$ for \methodtwo, same cost-budget rule for \methodthree's $\lambda$. The only thing that changes per row is the classifier (for \methodtwo) or the
route-value regressor head (for \methodthree).

We tried six model families. The tree-based ones are sklearn GBM, XGBoost, LightGBM, and CatBoost, all with the same hyperparameters (100 trees, max depth 3, learning rate 0.1, and subsample 0.8). The linear baseline is logistic regression for \methodtwo and Ridge Regression for \methodthree. The MLP baseline is a single hidden layer of 32 ReLU units, the same architecture for both routers. For the linear models and the MLP, we standardized the features first.
Tree models are scale-invariant, so we did not.

The take-away is the same as the main text. For \methodtwo, the routed F1 spans only $0.011$ across all six classifiers ($0.520$ to $0.531$). For \methodthree, it spans $0.016$ ($0.546$ to $0.562$). The deployed sklearn GBM is not strictly the best in either case; every alternative falls within the $95\%$ bootstrap CI on F1.
What drives RASER's performance is not the choice of model but the six features.

The full table also makes one thing visible that the main text does not: the linear and shallow models do not really learn the middle route. In \methodthree, Ridge sends only $5\%$ of questions to \pruneroute, and the MLP only $6\%$; the four tree-based regressors send $7$--$12\%$. Ridge and the MLP flatten the middle option, defaulting to either \stoproute or \iterroute. The tree models learn to use \pruneroute as a real cheap middle choice, and that is where
the extra $0.013$--$0.016$ F1 comes from.

\begin{table*}[t]
\centering
\footnotesize
\setlength{\tabcolsep}{4pt}
\renewcommand{\arraystretch}{0.96}
\begin{tabular}{llccc}
\toprule
Router & Model & \fone & Avg. tokens & Route mix (\%) \\
\midrule
\multirow{6}{*}{\methodtwo}
& \texttt{sklearn} GBM & 0.520 & 1514 & 86/14 \\
& Logistic Regression  & 0.531 & 1549 & 85/15 \\
& MLP-32               & 0.530 & 1597 & 82/18 \\
& XGBoost              & 0.526 & 1575 & 83/17 \\
& LightGBM             & 0.526 & 1607 & 82/18 \\
& CatBoost             & 0.529 & 1555 & 84/16 \\
\midrule
\multirow{6}{*}{\methodthree}
& \texttt{sklearn} GBM & 0.562 & 2157 & 64/12/24 \\
& Ridge Regression     & 0.546 & 1844 & 76/5/18 \\
& MLP-32               & 0.559 & 1908 & 74/6/19 \\
& XGBoost              & 0.562 & 2094 & 67/10/23 \\
& LightGBM             & 0.558 & 1937 & 73/9/18 \\
& CatBoost             & 0.562 & 1989 & 72/7/21 \\
\bottomrule
\end{tabular}
\caption{
Classifier ablation. All models are evaluated on the same setting over six LLMs and three datasets. For \methodtwo, different binary classifiers, produce a narrow \fone range of $0.011$. For \methodthree, different route-value regressors, a similarly narrow \fone range of $0.016$. Thus, the main conclusion is not tied to a particular implementation of the router.
We use the \texttt{sklearn} gradient boosting model because it is simple, standard, and reproducible, not because it is uniquely the best. S/P/I denotes the percentage of questions routed to \stoproute, \pruneroute, and \iterroute.
}
\label{tab:model_ablation}
\end{table*}

\section{Threshold and cost-budget sensitivity}
\label{app:sensitivity}

RASER has two settings the operator can adjust:

\textbf{\methodtwo's threshold $\theta$} controls how confident the router has to be before escalating to \pruneroute. Lower $\theta$ means the router escalates more often; higher $\theta$ means it stays
close to one-shot RAG.

\textbf{\methodthree's cost-budget fraction} controls how much the router is allowed to spend. We set this as a percentage of always-\iterroute's token cost (on training data); the actual $\lambda$ in Eq.~\ref{eq:route_argmax} is then derived to satisfy this budget.

In the paper, we use $\theta = 0.20$ and cost-budget $= 0.60$. This appendix sweeps both settings to show our choices are not cherry-picked: small changes give small differences. Everything else is held fixed (the same six features, the same 5-fold dataset cross-validation, and the same \texttt{sklearn} GBM head, pooled across all six LLMs and three datasets).

\paragraph{Threshold $\theta$ sweep for \methodtwo.} Table~\ref{tab:theta_sweep} shows what happens as $\theta$ moves
from $0.10$ (escalate aggressively) to $0.30$ (stay close to one-shot RAG). Both \fone and tokens move smoothly—no cliff, no hidden sweet spot. Across all five settings, \fone spans only $0.014$ and tokens span $349$. The deployed $\theta = 0.20$ is not the highest-\fone setting; $\theta = 0.10$ would gain $0.010$ more \fone but cost $242$ extra tokens per question. We picked $0.20$ because it gives the smallest token cost while staying within $0.010$ \fone of the maximum.

\begin{table}[h]
\centering\small
\setlength{\tabcolsep}{6pt}
\begin{tabular}{cccc}
\toprule
$\theta$ & \fone & Avg tokens & Escalation \% \\
\midrule
0.10                    & 0.530 & 1{,}756 & 24.0\% \\
0.15                    & 0.525 & 1{,}604 & 17.9\% \\
\textbf{0.20 (deployed)} & \textbf{0.520} & \textbf{1{,}514} & \textbf{14.1\%} \\
0.25                    & 0.518 & 1{,}455 & 11.6\% \\
0.30                    & 0.516 & 1{,}407 &  9.7\% \\
\bottomrule
\end{tabular}
\caption{\methodtwo threshold sensitivity. Across the five $\theta$ values the routed \fone moves by $0.014$ total and the token cost moves by $349$. The deployed $\theta = 0.20$ is not the \fone maximum but is within $0.010$ of it while spending $242$ fewer tokens per question.}
\label{tab:theta_sweep}
\end{table}

\paragraph{Cost-budget sweep for \methodthree.}
The raw $\lambda$ in the cost-aware argmax (Eq.~\ref{eq:route_argmax})
is hard to interpret on its own: ``$\lambda = 5 \times 10^{-5}$''
tells you nothing about how much the router will spend. We use a
plain-English setting instead: \emph{spend at most $X\%$ of
always-\iterroute's tokens on the training fold}. We then pick the
largest $\lambda$ that meets this limit, automatically per (LLM,
dataset) cell.

Table~\ref{tab:lambda_sweep} sweeps the limit from $0.33$ to
$1.00$. The two ends are sanity checks: at $0.33$ the router is
forced to be cheap and collapses to \stoproute ($99\%$ of questions
stop), so \methodthree behaves like one-shot RAG; at $1.00$ it can
freely use \iterroute on almost half the questions and gets the
highest \fone ($0.583$). The deployed $0.60$ sits at the knee of
the curve: moving the budget from $0.33$ to $0.60$ buys $+0.058$
\fone; moving from $0.60$ to $1.00$ buys only another $+0.021$ but
costs $50\%$ more tokens. This is the same trade-off as the Pareto
curve in Figure~\ref{fig:pareto_3route}, but with the setting expressed
as a token-spend percentage instead of a raw $\lambda$ number.

\begin{table}[h]
\centering\small
\setlength{\tabcolsep}{4pt}
\begin{tabular}{ccccc}
\toprule
Budget & \fone & Avg tokens & Route rate \\
\midrule
0.33                    & 0.504 & 1{,}190 & $99/1/0$ \\
0.50                    & 0.525 & 1{,}653 & $82/7/11$ \\
\textbf{0.60 (deployed)} & \textbf{0.562} & \textbf{2{,}157} & $\mathbf{64/12/24}$ \\
0.75                    & 0.578 & 2{,}665 & $44/19/37$ \\
1.00                    & 0.583 & 3{,}141 & $25/28/47$ \\
\bottomrule
\end{tabular}
\caption{\methodthree cost-budget sensitivity. Increasing the budget
from $0.33$ to $1.00$ traces the same Pareto curve as
Figure~\ref{fig:pareto_3route}, but parameterized by interpretable
fractions instead of $\lambda$ values. The deployed $0.60$ is
balanced: most of the \fone gain (above always-\stoproute at $0.33$)
is bought by the first $30$ percentage points of budget;
the last $40$ \% points add little \fone but spend a lot
more tokens.}
\label{tab:lambda_sweep}
\end{table}

\end{document}